\documentclass[letterpaper]{article} % DO NOT CHANGE THIS
\usepackage[preprint]{aaai2027}  % DO NOT CHANGE THIS
\usepackage[hyphens]{url}  % DO NOT CHANGE THIS
\usepackage{graphicx} % DO NOT CHANGE THIS
\usepackage{natbib}  % DO NOT CHANGE THIS AND DO NOT ADD ANY OPTIONS TO IT
\usepackage{caption} % DO NOT CHANGE THIS AND DO NOT ADD ANY OPTIONS TO IT
\usepackage{amssymb}
\usepackage{amsmath}
\usepackage{adjustbox}
\usepackage{multirow}
\usepackage{graphicx}
\usepackage{booktabs}
\usepackage{multirow}
\usepackage[table]{xcolor}
\usepackage{xcolor}
\usepackage{float}
\usepackage{booktabs,multirow,tabularx,array,makecell,colortbl,xcolor}
\usepackage{array}

\newcolumntype{Y}{>{\centering\arraybackslash}X}
\newcolumntype{Z}{>{\columncolor{green!10}\centering\arraybackslash}X}

\usepackage{algorithm}
\usepackage{algorithmic}

\usepackage{newfloat}
\usepackage{listings}
\DeclareCaptionStyle{ruled}{labelfont=normalfont,labelsep=colon,strut=off} % DO NOT CHANGE THIS
\floatstyle{ruled}
\newfloat{listing}{tb}{lst}{}
\floatname{listing}{Listing}

\usepackage{booktabs}

\title{GLOBE: Trajectory-Aligned Gradient Matching with Structured SparseOptimization for Coreset Selection}
\author {
    Hetian Liu\textsuperscript{\rm 1, \rm 3}\equalcontrib,
    Jin Cui\textsuperscript{\rm 1}\equalcontrib,
    Mengcheng Shi\textsuperscript{\rm 2},
    Yanbin Hu\textsuperscript{\rm 1,\rm 3},
    Xinyue Long\textsuperscript{\rm 1,\rm 3},
    Boran Zhao\textsuperscript{\rm 1,\rm 3}\corresponding,
    Pengju Ren\textsuperscript{\rm 1}
}
\affiliations {
    \textsuperscript{\rm 1}State Key Laboratory of Human-Machine Hybrid Augmented Intelligence,\\
    and Institute of Artificial Intelligence and Robotics, Xi'an Jiaotong University\\
    \textsuperscript{\rm 2}School of Computer Science and Technology, Xi'an Jiaotong University \\
    \textsuperscript{\rm 3}School of Software Engineering, Xi'an Jiaotong University \\
    andycui@stu.xjtu.edu.cn
}

\begin{document}

\maketitle

\begin{abstract}
On-device training of deep neural networks is fundamentally constrained by the computational and memory costs of large-scale datasets. Coreset selection offers a practical solution by retaining only a compact subset of real training samples. However, existing gradient-based methods commonly rely on gradients computed at a single model snapshot and employ greedy or pursuit-based selection procedures, limiting their ability to capture evolving optimization dynamics and handle strongly correlated samples. We propose GLOBE (Gradient Local–Balanced Extraction), a trajectory-aligned coreset selection framework that formulates sample selection as a globally optimized sparse weighting problem. GLOBE represents each sample by a gradient trajectory constructed across multiple training checkpoints, thereby capturing its influence throughout different stages of optimization. To preserve the training behavior of the full dataset, we introduce a multi-order matching objective that jointly aligns the first-order mean and projected uncentered second-order moments of gradient trajectories. GLOBE further combines Group LASSO, Elastic Net regularization, and nonnegative budget constraints to induce group- and sample-level sparsity while stabilizing the weights of correlated trajectories. Finally, class-balanced Top-\(K\) selection maintains adequate category coverage under limited sampling budgets. Experiments across six benchmarks and five evaluation architectures demonstrate that GLOBE consistently outperforms existing coreset selection methods in downstream test accuracy, particularly at low retention ratios. These results highlight the effectiveness of combining dynamic gradient information, multi-order distribution matching, and structured sparsity for data-efficient learning.

\end{abstract}


\section{Introduction}

Deep neural networks (DNNs) have made remarkable progress recently and have achieved or even surpassed human-level performance on specific tasks in computer vision~\cite{cv1, cv2, cv3}, natural language processing~\cite{attention_nlp1, bert_nlp3, gpt3_nlp2}, and scientific computing~\cite{sc1, sc2}. Meanwhile, with the rapid development of edge intelligence applications such as the Internet of Things and autonomous driving, together with growing concerns over communication latency and data privacy, there is an increasing demand for training and updating neural networks directly on edge devices. However, the success of DNNs relies heavily on large-scale training data~\cite{train_heave,coco}, whereas edge devices are typically subject to strict constraints on computational resources, storage capacity, and energy budgets, making it impractical to train models directly on the full dataset. Therefore, reducing the data scale while preserving its training utility has become an important challenge for on-device learning. To address this issue, various dataset compression techniques have been proposed~\cite{dc_survey}, among which dataset distillation~\cite{datasynthesis_1,datasynthesis_2,datasynthesis_3, datasynthesis_4} and coreset selection~\cite{coreset1,coreset3,GM_coreset2} are two representative paradigms. Dataset distillation synthesizes a set of training samples to approximate the training effect of the full dataset, whereas coreset selection directly identifies a representative subset from the original data. In comparison, coreset selection achieves a favorable balance between computational efficiency and data fidelity, making it suitable for resource-constrained edge deployment.

\begin{figure}[t]
    \centering
    \includegraphics[width=1\linewidth]{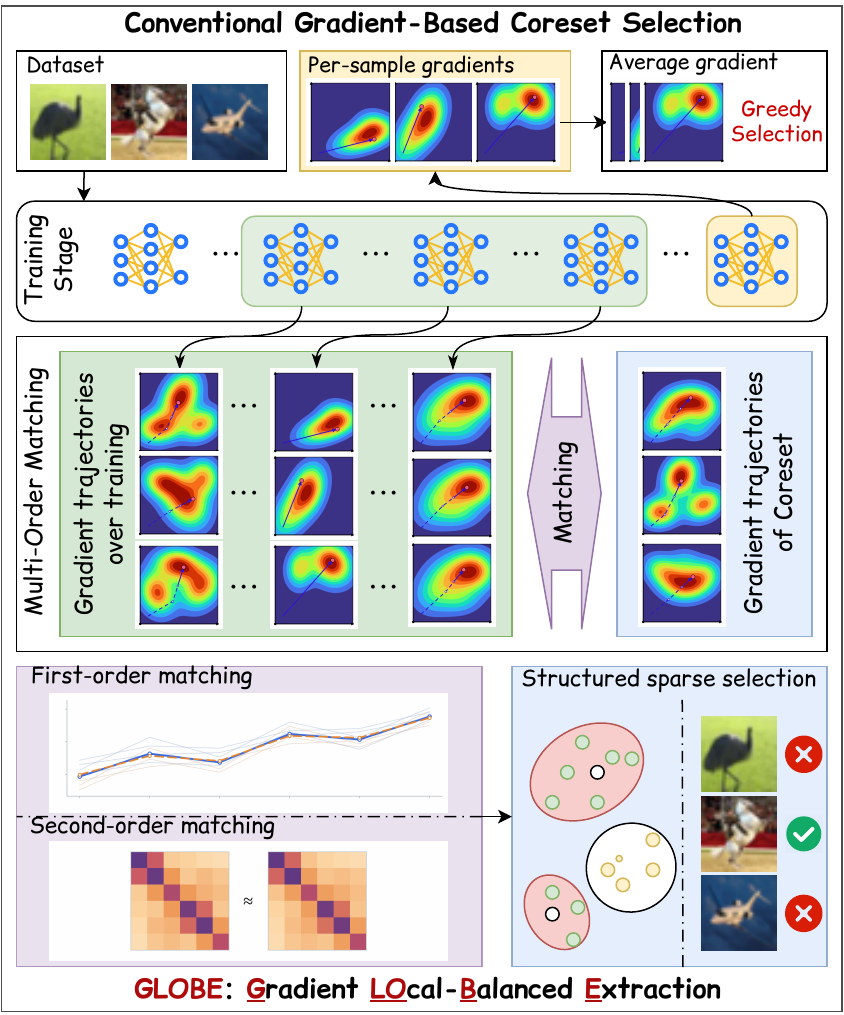}
    \caption{\textbf{Comparison between conventional gradient-based coreset selection and GLOBE.} Conventional methods select samples using gradients from a single model snapshot, whereas GLOBE models multi-checkpoint gradient trajectories, matches their first- and second-order statistics, and applies structured sparse optimization to obtain a compact and representative coreset.}
    \label{fig_intro}
\end{figure}

Coreset selection methods based on gradient matching aim to construct a small subset whose gradients approximate those of the full dataset, thereby identifying the most representative training samples. However, as shown in Figure~\ref{fig_intro}, representative methods such as CRAIG~\cite{craig} and GradMatch~\cite{GM_coreset2} mainly rely on single-step gradients computed at a fixed model snapshot, which limits their ability to capture the temporal evolution of training dynamics. In addition, many existing methods~\cite{craig,GM_coreset2,glister} adopt greedy optimization procedures, such as facility-location-based greedy selection or Orthogonal Matching Pursuit. These methods typically make locally myopic decisions and may suffer from performance degradation when strong correlations exist among samples. Moreover, to reduce the computational cost of per-sample gradient extraction, existing methods commonly employ lightweight proxy networks. However, structural discrepancies between the proxy and target models may introduce additional gradient bias, which can further affect the quality of the selected coreset.

In this work, we revisit gradient-based coreset selection from both dynamic and distributional perspectives. We first introduce gradient trajectories, which aggregate per-sample gradients across multiple checkpoints during training to jointly characterize the temporal evolution and multi-stage geometry of the optimization path. To more accurately approximate the gradient distribution of the full dataset, we construct a two-level matching objective that jointly aligns the first-order statistics, namely the mean, and the second-order statistics, namely the covariance, of the gradient trajectories. This design captures both the overall gradient direction and the correlation structure among sample gradients. We then formulate coreset selection as a sample-weight optimization problem with structured sparsity constraints. Specifically, Group LASSO induces group-level sparsity over similarity-based groups, while Elastic Net promotes sample-level sparsity and stabilizes the weights of correlated samples within the retained groups. Finally, we introduce a lightweight teacher-to-proxy alignment mechanism to ensure that gradient trajectories computed using a compact proxy model remain faithful to the gradient behavior of the target model.

Based on these designs, we propose GLOBE (Gradient Local--Balanced Extraction). Our main contributions are:
\begin{itemize}
    \item We introduce multi-checkpoint gradient trajectories and a multi-order matching objective that preserves the first-order mean and projected second-order moments of the full-data trajectory distribution.

    \item We formulate coreset selection as a globally optimized sparse weighting problem, combining Group LASSO and Elastic Net to induce group- and sample-level sparsity while stabilizing correlated sample weights.

    \item Experiments on six benchmarks and five architectures show that GLOBE consistently outperforms existing methods, particularly at low retention ratios.
\end{itemize}

\section{Related Works}

\subsection{Dataset Compression}
Dataset compression methods can be broadly categorized into dataset distillation~\cite{datasynthesis_1,datasynthesis_2,datasynthesis_3,ds_parameter} and coreset selection~\cite{coreset1,coreset3,GM_coreset2}. Dataset distillation optimizes a compact set of synthetic samples so that a model trained on them achieves performance comparable to that obtained using the full dataset. Representative approaches include bilevel-optimization-based dataset distillation~\cite{bilevel}, gradient matching~\cite{gradientmatch_1, gradientmatch_2}, distribution matching~\cite{ds_parameter}, and training trajectory matching~\cite{trajectory_1,trajectory_2}. However, these methods typically require repeated model training and gradient backpropagation to optimize the synthetic samples, resulting in substantial computational overhead. Moreover, synthetic samples may suffer from limited interpretability and restricted generalization across model architectures. In contrast, coreset selection directly selects representative samples from the original dataset, thereby preserving the semantic content of real data. Existing methods mainly assess sample importance based on geometric coverage~\cite{geo1,geo2}, uncertainty~\cite{unc1,unc2}, or decision-boundary information~\cite{boundary1,boundary2}. Although coreset selection can preserve much of the information contained in the
full dataset, its effectiveness remains highly dependent on the selection criterion. Accurately characterizing the contribution of each sample therefore remains a key challenge.

\subsection{Gradient-Based Dataset Compression}

Gradient signals directly reflect how individual samples influence model parameter updates and have consequently been widely used in dataset compression. In dataset distillation, early methods optimized synthetic samples by matching the single-step gradients of real and synthetic data. Subsequent approaches extended single-step gradient matching by accumulating
errors over multiple parameter updates or performing sequential matching~\cite{multi_step1,multi_step2}, while another line of work directly matched longer training trajectories~\cite{traject_1,traject_2} to preserve more complete training dynamics. In coreset selection, GraNd~\cite{GraNd} estimates sample importance using gradient norms, CRAIG~\cite{craig} selects representative samples to approximate the full-data gradient, and GradMatch~\cite{GM_coreset2} employs Orthogonal Matching Pursuit to construct a weighted subset. However, existing gradient-based coreset selection methods typically rely on gradients computed from a single model at a fixed training snapshot, making it difficult to capture how sample contributions evolve across different stages of training. Moreover, they primarily focus on matching the mean gradient while paying limited attention to the correlation structure among sample gradients. Greedy selection and Orthogonal Matching Pursuit may also produce locally myopic and unstable solutions when candidate samples are highly correlated.

\subsection{Sparse Optimization for Coreset Selection}

Sparse optimization provides a natural framework for selecting a compact set of representative samples from a large candidate pool~\cite{sparse_rep1,sparse_rep2}. The Least Absolute Shrinkage and Selection Operator (LASSO) promotes sample-level sparsity through \(\ell_1\) regularization, driving many sample coefficients exactly to zero~\cite{lasso}. However, real-world datasets often contain samples with strongly correlated features or gradients, under which LASSO may exhibit unstable selection behavior and arbitrarily retain only a subset of correlated candidates~\cite{elastic_net,lasso_corr}. Elastic Net combines \(\ell_1\) and \(\ell_2\) regularization~\cite{elastic_net}. While the \(\ell_1\) term preserves sample-level sparsity, the \(\ell_2\) term encourages smoother coefficient shrinkage and improves stability among correlated samples. Group LASSO further imposes structured sparsity on predefined groups, allowing entire groups of related samples to be selected or discarded jointly~\cite{group_lasso}. In this work, we integrate these structured sparse regularization terms with gradient trajectory matching and formulate coreset selection as a unified weight-optimization problem for stable and representative sample selection.

\section{Method}
\label{sec:method}

\subsection{Problem Formulation}

Given a labeled dataset \(\mathcal{D}=\{(x_i,y_i)\}_{i=1}^{N},\) our goal is to construct a compact coreset \(\mathcal{S}\subset\mathcal{D}\) of size \(K\) that approximates the training dynamics of the full dataset.

Instead of selecting samples explicitly, we treat coreset construction as learning a sparse weight vector \(\omega\in\mathbb{R}_{\geq 0}^{N}.\) The coefficient \(\omega_i\) quantifies the contribution of sample \(i\) to reconstructing the full-data gradient trajectory statistics. We subsequently use the coefficient magnitudes to obtain a discrete coreset through class-balanced Top-\(K\) selection.

Let \(\theta\) denote the parameters of a lightweight proxy model, and let \(\mathcal{J}\) denote the index set of tracked layers. For a training sample \((x_i,y_i)\), we define its gradient at layer \(j\in\mathcal{J}\) as
\begin{equation}
    g_i^{(j)}(\theta)
    =
    \nabla_{\theta^{(j)}}
    \ell(x_i,y_i;\theta)
\end{equation}
Classical gradient matching constructs a weighted subset whose aggregate gradient approximates the mean gradient of the full dataset:
\begin{equation}
\min_{\substack{\mathcal{S}\subset\mathcal{D}}}
\sum_{j\in\mathcal{J}}
\alpha_j\left|\sum_{x_i\in\mathcal{S}}\omega_i g_i^{(j)}(\theta)-\frac{1}{N}\sum_{i=1}^{N}g_i^{(j)}(\theta)\right|^2,
\end{equation}
where \(\omega_i\) denotes the reconstruction weight of the selected sample and \(\alpha_j\) balances the contributions of different tracked layers. 

\begin{figure*}[t]
    \centering
    \includegraphics[width=1\linewidth]{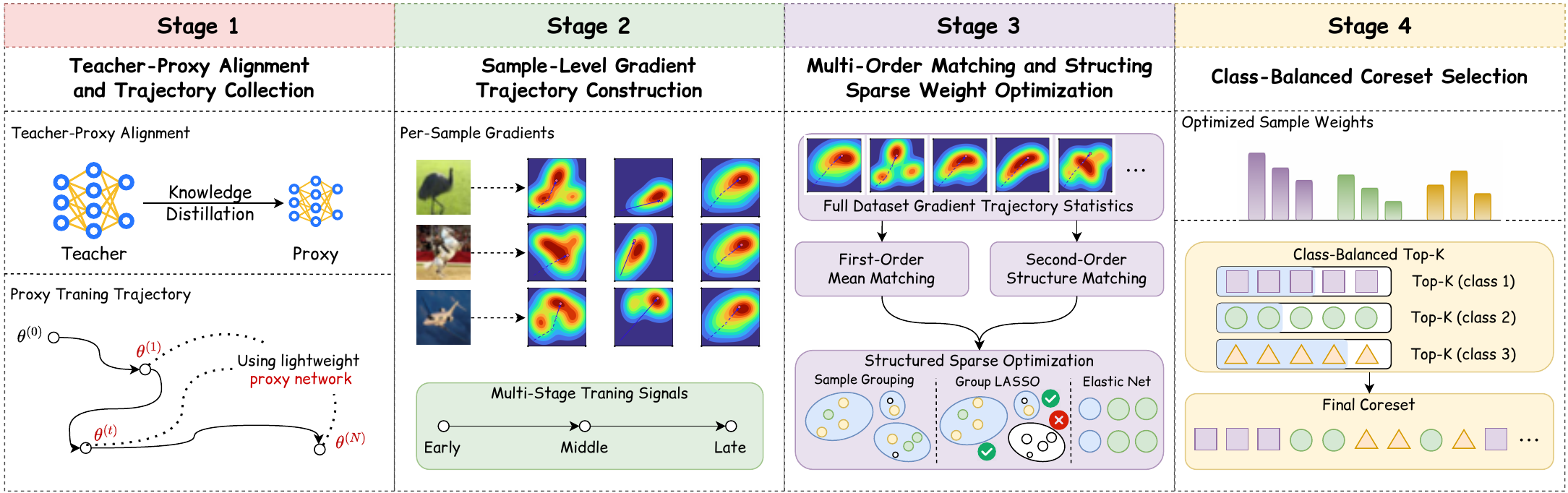}
    \caption{\textbf{Overview of the GLOBE framework.} GLOBE first aligns a lightweight proxy model with the teacher model and collects multiple training checkpoints. It then constructs sample-level gradient trajectories, matches the first- and second-order trajectory statistics of the full dataset, and learns sparse sample weights using Group LASSO and Elastic Net. Finally, class-balanced Top-\(K\) selection produces the coreset.}
    \label{fig_framework}
\end{figure*}

In contrast, we represent each sample not by a single gradient, but by its \emph{gradient trajectory} across multiple training checkpoints. Our objective is to approximate the full-data distribution of gradient trajectories by matching both first-order means and second-order statistics. The overall pipeline is illustrated in Figure~\ref{fig_framework}.

\subsection{Gradient Trajectory Construction}
\label{sec:trajectory}

\paragraph{Teacher-proxy alignment.}
We employ a lightweight proxy network to extract per-sample gradients across the training stage. To reduce the discrepancy between the proxy and a stronger target architecture, the proxy can be initialized through knowledge distillation from a teacher network.

Let \(z_s\) and \(z_t\) denote the proxy and teacher logits, respectively, and define
\begin{equation}
    p_s^{\tau}=\operatorname{softmax}(z_s/\tau),
    \qquad
    p_t^{\tau}
    =
    \operatorname{softmax}(z_t/\tau),
\end{equation}
where \(\tau\) is the distillation temperature. The distillation objective is
\begin{equation}
\mathcal{L}_{\mathrm{KD}}=\eta\tau^{2}D_{\mathrm{KL}}
\big(
    p_t^{\tau}\,\|\,p_s^{\tau}
\big)+(1-\eta)\operatorname{CE}(z_s,y),
\end{equation}
where \(\eta\in[0,1]\) controls the balance between soft-label distillation and hard-label supervision. This procedure aligns the output distributions of the teacher and proxy rather than explicitly aligning their intermediate features or gradients. This alignment of the output-level reduces the discrepancy between the two models while preserving the efficiency of proxy training, thereby providing a reliable basis for the subsequent construction of gradient trajectories.

\paragraph{Dynamic gradient trajectories.}
We train the proxy for a number of epochs and record \(T\) checkpoints, denoted by \(\Theta=\{\theta^{(1)},\ldots,\theta^{(T)}\}.\)
At each checkpoint, we compute layer-wise per-sample gradients over the selection pool. For sample \(i\) and tracked layer \(j\), its gradient trajectory is defined as
\begin{equation}
    G_i^{(j)}=\operatorname{concat}\left(g_i^{(j)}(\theta^{(1)}),\ldots,g_i^{(j)}(\theta^{(T)})\right)\in\mathbb{R}^{d_jT},
\end{equation}
where \(d_j\) is the number of parameters in layer \(j\). Thus, gradient trajectories are constructed separately for each tracked layer. Compared with single-step gradients, these trajectories capture both early-stage optimization signals (e.g., coarse decision boundaries) and late-stage refinements (e.g., classspecific discrimination), providing a richer descriptor of sample importance.

To prepare the per-sample trajectories for full-data matching, we first aggregate the trajectories of all samples within each tracked layer into a layer-wise gradient block.
For layer \(j\), we stack all sample trajectories row-wise:
\begin{equation}
G^{(j)}
=
\left[
\begin{array}{c}
(G_1^{(j)})^{\top}\\
\vdots\\
(G_N^{(j)})^{\top}
\end{array}
\right]
\in\mathbb{R}^{N\times d_jT}
\end{equation}
The layer-wise systems are then vertically concatenated:
\begin{equation}
A=
\left[
\begin{array}{c}
\sqrt{\alpha_{j_1}}\,(G^{(j_1)})^{\top}\\
\vdots\\
\sqrt{\alpha_{j_{|\mathcal{J}|}}}\,
(G^{(j_{|\mathcal{J}|})})^{\top}
\end{array}
\right]
\in
\mathbb{R}^{\sum_j d_jT\times N}
\end{equation}
The corresponding full-data target is obtained by averaging the gradient trajectories of all samples. To balance the contributions of different layers, each layer is weighted inversely proportional to its trajectory dimensionality, preventing high-dimensional layers from dominating the matching objective.

\subsection{Multi-Order Distribution Matching}

Gradient trajectories provide a high-dimensional representation of how each sample influences learning across the entire optimization process. To obtain a coreset that faithfully preserves these signals, we go beyond single-point gradient matching and introduce a two-level distribution-matching objective that aligns both the \emph{mean} and the \emph{covariance} structure of gradient trajectories.

\paragraph{First-order matching.}
The first-order matching objective is
\begin{equation}
    \mathcal{L}_{\mathrm{mean}}(\omega)
    =
    \|A\omega-b\|_2^2.
\end{equation}
The vector \(A\omega\) is a weighted reconstruction of the full-data mean trajectory in the scaled and stacked coordinate system. With the sparse regularization introduced later, it is interpreted as a sparse weighted reconstruction rather than a normalized weighted mean.

\paragraph{Second-order matching.}
First-order alignment alone fails to capture higher-order geometry of gradient trajectories, such as variability and anisotropy across samples. To characterize correlations among gradient trajectory dimensions without explicitly constructing full second-order matrices in \(\mathbb{R}^{d_jT}\), we apply an independent random projection to each second-order layer \(j\in\mathcal{J}_2\), where \(\mathcal{J}_2\subseteq\mathcal{J}\).
Let
\begin{equation}
    R_j
    \in
    \mathbb{R}^{d_jT\times m}
\end{equation}
be the random projection matrix for layer \(j\). The projected trajectory and
its uncentered outer product are
\begin{equation}
    v_i^{(j)}
    =
    R_j^{\top}G_i^{(j)}
    \in
    \mathbb{R}^{m},
    \qquad
    Q_i^{(j)}
    =
    v_i^{(j)}(v_i^{(j)})^{\top}
    \in
    \mathbb{R}^{m\times m}.
\end{equation}

The full-data uncentered second moment at layer \(j\) is
\begin{equation}
    \overline{Q}^{(j)}
    =
    \frac{1}{N}
    \sum_{i=1}^{N}
    Q_i^{(j)},
\end{equation}
whereas the weighted reconstruction is
\begin{equation}
    \widehat{Q}^{(j)}(\omega)
    =
    \sum_{i=1}^{N}
    \omega_i Q_i^{(j)}.
\end{equation}
We emphasize that these quantities are uncentered second moments rather than
centered covariance matrices. The second-order matching objective is
\begin{equation}
    \mathcal{L}_{\mathrm{sec}}(\omega)
    =
    \sum_{j\in\mathcal{J}_2}
    \beta_j
    \left\|
        \widehat{Q}^{(j)}(\omega)
        -
        \overline{Q}^{(j)}
    \right\|_F^2,
\end{equation}
where \(\beta_j\geq0\) controls the contribution of each layer.

For implementation, the matrices
\(
Q_i^{(j)}
\)
are vectorized and stacked into a linear system. Let \(B\) denote the
resulting stacked operator and let \(c\) denote the corresponding stacked
full-data target. The second-order loss can equivalently be written as
\begin{equation}
    \mathcal{L}_{\mathrm{sec}}(\omega)
    =
    \|B\omega-c\|_2^2.
\end{equation}

\paragraph{Combined matching objective.}
The complete gradient trajectory matching loss is
\begin{equation}
    \mathcal{L}_{\mathrm{match}}(\omega)
    =
    \gamma_1
    \mathcal{L}_{\mathrm{mean}}(\omega)
    +
    \gamma_2
    \mathcal{L}_{\mathrm{sec}}(\omega),
\end{equation}
which ensures that the coreset matches not only the average gradient trajectory but also the underlying covariance
structure governing optimization dynamics. This two-level matching serves as the foundation upon which we introduce structured sparsity to obtain a compact and semantically meaningful coreset.

\subsection{Structured Sparse Optimization}
\label{sec:structured-sparse}

While distribution matching provides a principled objective for preserving training dynamics, optimizing over all \(N\) samples is infeasible for large datasets. To obtain a compact coreset, we impose structured sparsity on the weight vector \(\omega\), combining Group LASSO and Elastic Net regularization under a simplex constraint. This design enables both semantic group-level selection and stable fine-grained sparsification.

\paragraph{Sample grouping.}
Real-world datasets often contain semantically coherent samples that exhibit highly correlated gradient trajectories. Direct sparsification at the individual-sample level may therefore produce unstable or arbitrarily sparse solutions. To capture this structure, we extract penultimate-layer proxy features and perform class-wise \(k\)-means clustering. This produces a collection of non-overlapping sample groups \(\mathcal{G}=\{G_m\}_{m=1}^{M},\) where each \(G_m\) contains samples with similar representations.

\paragraph{Group-level sparsity.}
To encourage selection at the group level, we apply the Group LASSO regularizer
\begin{equation}
    \mathcal{R}_{\mathrm{group}}(\omega)
    =
    \lambda_{\mathrm{g}}
    \sum_{m=1}^{M}
    \|\omega_{G_m}\|_2,
\end{equation}
This term drives entire groups toward zero simultaneously, effectively pruning unimportant semantic clusters and reducing the candidate space before individual sample-level sparsification.

\paragraph{Sample-level sparsity.}
Within groups that remain active, individual trajectories may still be highly correlated. Relying solely on an $\ell_1$ penalty may result in unstable solutions in such regimes. To address this, we adopt an Elastic Net penalty:

\begin{equation}
    \mathcal{R}_{\mathrm{EN}}(\omega)
    =
    \lambda_1\|\omega\|_1
    +
    \lambda_2\|\omega\|_2^2.
\end{equation}
The \(\ell_1\) term encourages sample-level sparsity, while the \(\ell_2\)
term penalizes excessively concentrated coefficient assignments.

\paragraph{Overall objective.}
The coefficient vector is obtained by solving

\begin{equation}
\label{eq:overall-objective}
\begin{aligned}
\min_{\omega \geq 0}\quad
& \mathcal{L}_{\mathrm{match}}(\omega)
+ \mathcal{R}_{\mathrm{group}}(\omega)
+ \mathcal{R}_{\mathrm{EN}}(\omega) \\
={}\quad
& \|A\omega-b\|_2^2
+ \|B\omega-c\|_2^2 \\
& + \lambda_{\mathrm{g}}
\sum_{m=1}^{M}
a_m\|\omega_{G_m}\|_2 
 + \lambda_1\|\omega\|_1
+ \lambda_2\|\omega\|_2^2 .
\end{aligned}
\end{equation}

This formulation jointly enforces (i) first-order and second-order trajectory matching, (ii) semantic group-level sparsity, (iii) robust individual-level shrinkage, and (iv) probabilistic interpretability, providing a principled and stable mechanism for constructing compact, distribution-preserving coresets. We use a proximal-gradient-style solver. Each iteration consists of a smooth gradient step on the matching and \(\ell_2\) terms, followed by group \(\ell_2\) shrinkage, sample-wise \(\ell_1\) shrinkage, and Euclidean projection onto the selected feasible set.

\paragraph{From continuous coefficients to a discrete coreset.}
The optimized coefficients are continuous reconstruction coefficients rather
than a discrete subset. To construct a coreset of size \(K\), we assign a
budget \(K_r\) to each class, where
\begin{equation}
    \sum_{r=1}^{C}K_r=K.
\end{equation}
The equal-budget setting uses approximately \(K/C\) samples per class,
whereas an optional dynamic-budget strategy assigns \(K_r\) using
class-level coefficient statistics.

For each class, we rank the samples according to their optimized coefficients
and select the \(K_r\) largest values. The final coreset is
\begin{equation}
    \mathcal{S}
    =
    \bigcup_{r=1}^{C}
    \operatorname{TopK}
    \left(
        \{\omega_i:i\in\mathcal{I}_r\},
        K_r
    \right).
\end{equation}
The selected coreset is therefore not necessarily identical to the positive
support
\(
\{i:\omega_i>0\}.
\)
If a class contains fewer than \(K_r\) positive coefficients, the class-wise
Top-\(K_r\) operation necessarily includes zero or near-zero coefficient
samples. Finally, the selected pool indices are mapped back to the
corresponding indices in the original dataset.

\newcommand{\mr}[1]{\multirow{3}{*}{#1}}
\newcommand{\ds}[1]{\mr{\shortstack{#1}}}
\newcommand{\best}[1]{\textbf{#1}}

\begin{table*}[t]
\centering
\caption{Test accuracy (\%) on six image classification benchmarks under different data retention ratios. Results are averaged over five evaluation architectures. ``Full'' denotes training on the complete dataset. Abbreviations: NMS.\ = Near Memory Sampling~\cite{nms}, Herd.\ = Herding~\cite{herding}, k-Cen.\ = \(k\)-Center~\cite{geo1}, Ent.\ = Entropy~\cite{entropy}, Forg.\ = Forgetting~\cite{forgetting}, DF = DeepFool~\cite{boundary1}, CRA.\ = CRAIG~\cite{craig}, GM = GradMatch~\cite{GM_coreset2}, and GLIS.\ = GLISTER~\cite{glister}. Bold values indicate the best result for each setting.}

\label{tab:main_results}
\vspace{-1mm}
\scriptsize
\setlength{\tabcolsep}{3pt}         % 只控制列间距
\setlength{\aboverulesep}{0.4ex}   % 横线上方间距
\setlength{\belowrulesep}{0.4ex}   % 横线下方间距
\renewcommand{\arraystretch}{0.8}   % 控制行间距

\resizebox{\textwidth}{!}{%
\begin{tabular}{cc|*{14}{c}|c|>{\columncolor{green!10}}c}
\toprule
Dataset & Ratio & NMS & Herd. & k-Cen. & Ent. & Forg. & GraNd & CAL
& DF & CRA. & GM & GLIS. & FL & GC & DQ & Full & \best{GLOBE} \\
\midrule

\ds{CIFAR-\\10}
& 10\% & 85.33 & 71.02 & 77.01 & 56.28 & 68.71 & 44.71 & 81.32 & 59.07 & 61.21 & 60.10 & 56.86 & 77.83 & 84.13 & 84.82 & \mr{94.76} & \best{86.56} \\
& 20\% & 88.47 & 71.95 & 82.99 & 68.55 & 79.83 & 61.81 & 81.44 & 74.64 & 69.37 & 67.64 & 68.67 & 82.34 & 84.68 & 87.43 & & \best{89.89} \\
& 30\% & 90.80 & 77.88 & 88.21 & 81.32 & 86.47 & 74.48 & 85.24 & 80.84 & 80.85 & 78.51 & 78.22 & 85.99 & 88.31 & 90.79 & & \best{92.91} \\
\midrule

\ds{CINIC-\\10}
& 10\% & 60.96 & 29.09 & 57.81 & 38.22 & 62.26 & 24.78 & 56.29 & 46.24 & 53.28 & 45.53 & 45.51 & 61.99 & 59.17 & 62.88 & \mr{81.90} & \best{64.75} \\
& 20\% & 69.74 & 42.17 & 72.46 & 57.53 & 70.85 & 47.16 & 63.08 & 62.12 & 66.47 & 55.50 & 55.63 & 72.64 & 68.55 & 71.16 & & \best{73.23} \\
& 30\% & 75.60 & 51.76 & 77.68 & 71.00 & 75.82 & 65.25 & 68.70 & 71.28 & 70.99 & 67.63 & 67.11 & 76.09 & 73.08 & 74.73 & & \best{78.43} \\
\midrule

\mr{SVHN}
& 10\% & 90.56 & 61.21 & 81.73 & 69.67 & 77.18 & 50.27 & 80.73 & 66.50 & 65.11 & 61.57 & 58.09 & 80.35 & 87.83 & 89.91 & \mr{96.38} & \best{91.40} \\
& 20\% & 92.44 & 86.14 & 89.80 & 89.89 & 91.41 & 85.86 & 86.93 & 87.38 & 89.49 & 85.91 & 85.82 & 89.62 & 90.67 & 91.65 & & \best{93.13} \\
& 30\% & 93.13 & 92.36 & 93.21 & 93.39 & 93.61 & 92.46 & 89.67 & 92.39 & 92.88 & 91.83 & 92.02 & 92.74 & 92.31 & 92.72 & & \best{94.53} \\
\midrule

\ds{ImageNet-\\100}
& 10\% & 55.62 & 40.68 & 41.55 & 26.24 & 48.37 & 19.92 & 56.53 & 33.91 & 36.74 & 32.92 & 31.50 & 53.76 & 58.61 & 56.06 & \mr{81.13} & \best{60.69} \\
& 20\% & 63.89 & 50.24 & 53.81 & 39.61 & 58.44 & 28.03 & 63.32 & 43.98 & 43.31 & 43.74 & 40.94 & 58.48 & 65.04 & 60.27 & & \best{68.97} \\
& 30\% & 69.13 & 54.99 & 61.61 & 49.85 & 66.06 & 40.33 & 68.24 & 53.34 & 53.17 & 48.57 & 50.63 & 63.46 & 69.52 & 64.87 & & \best{72.08} \\
\midrule

\ds{ImageNet-\\1K}
& 10\% & 49.73 & 31.71 & 31.61 & 21.83 & 38.49 & 18.84 & 39.92 & 30.12 & 32.86 & 35.92 & 34.71 & 41.68 & 39.57 & 38.22 & \mr{69.22} & \best{50.32} \\
& 20\% & 53.22 & 41.54 & 44.24 & 30.17 & 50.07 & 21.77 & 49.55 & 42.55 & 40.39 & 41.75 & 40.83 & 46.73 & 48.89 & 45.56 & & \best{54.79} \\
& 30\% & 58.98 & 53.65 & 57.48 & 45.11 & 58.00 & 34.64 & 56.98 & 51.43 & 48.99 & 54.40 & 54.26 & 55.33 & 55.51 & 48.52 & & \best{59.85} \\
\midrule

\ds{CIFAR-\\100}
& 10\% & 48.02 & 25.95 & 33.77 & 26.16 & 47.41 & 34.49 & 50.62 & 34.05 & 45.62 & 44.55 & 43.80 & 49.04 & 46.73 & 45.23 & \mr{73.54} & \best{50.76} \\
& 20\% & 57.38 & 33.05 & 44.58 & 36.69 & 56.55 & 44.40 & 57.06 & 47.66 & 52.57 & 53.28 & 52.50 & 52.00 & 56.89 & 56.80 & & \best{59.44} \\
& 30\% & 64.04 & 49.45 & 48.78 & 37.95 & 59.41 & 45.52 & 62.27 & 53.01 & 55.02 & 57.58 & 58.11 & 48.85 & 59.21 & 60.24 & & \best{65.87} \\

\bottomrule
\end{tabular}}
\vspace{-3mm}
\end{table*}

\section{Experiment and Aanlysis}

\label{sec:experiment}
\subsection{Experiment Setups}
\noindent\textbf{Datasets.} 
We evaluate GLOBE on six image classification benchmarks: CIFAR-10, CIFAR-100~\cite{CIFAR-10_100}, CINIC-10~\cite{cinic}, SVHN~\cite{SVHN}, ImageNet-100, and ImageNet-1K~\cite{imagenet}. To assess its effectiveness across datasets of different scales and complexities, we organize them into two groups. The first group consists of CIFAR-10, CINIC-10, and SVHN, each containing 10 classes with (32$\times$32) images. These datasets cover natural-object recognition, cross-dataset distribution variation, and digit classification. The second group includes CIFAR-100, ImageNet-100, and ImageNet-1K, which provide increasingly challenging settings in terms of category scale and image resolution. CIFAR-100 contains 100 classes with (32$\times$32) images, ImageNet-100 contains 100 classes with (224$\times$224) images, and ImageNet-1K further expands the evaluation to 1,000 classes at the same resolution.

\noindent\textbf{Models and Baseline.}
To evaluate the cross-architecture generalization of GLOBE, we consider five representative network architectures, including standard convolutional networks (ResNet18 and ResNet50~\cite{resnet}), lightweight models designed for efficient deployment (ShuffleNetV2~\cite{shufflenet} and MobileNetV2~\cite{mobilenetv2}), and a vision Transformer (ViT)~\cite{vit}. We evaluate data retention ratios of 10\%, 20\%, and 30\% to examine performance under different compression levels. We compare GLOBE against a comprehensive set of baselines, categorized as shown in Table~\ref{tab:main_results}.

% Unless otherwise specified, all models are trained for 200 epochs using SGD with a momentum of 0.9 and a weight decay of (5\times10^{-4}). The batch size is set to 256 for datasets with (32\times32) images and 64 for higher-resolution datasets with (224\times224) or (256\times256) images.

\noindent\textbf{Implementation.}
The hardware server used in our experiments is equipped with an AMD $EPYC^{TM}$ 7H12 64-Core Processor, four NVIDIA RTX 6000 Ada GPUs, and 512 GB of system memory.

\subsection{Results and Analysis}
\noindent\textbf{Overall Performance.} 
We validated GLOBE across datasets of different scales and data-retention ratios. Table~\ref{tab:main_results} summarizes the results on six image-classification benchmarks at coreset ratios of 10\%, 20\%, and 30\%. GLOBE achieves the highest accuracy in all 18 experimental settings, consistently outperforming classical geometric selection, score-based pruning, decision-boundary methods, gradient-matching approaches, and bilevel-optimization baselines. The advantage of GLOBE remains evident under aggressive compression. At the 10\% retention ratio, GLOBE improves over the strongest baseline by 1.23 percentage points on CIFAR-10 (86.56\% vs.\ 85.33\%), 1.87 points on CINIC-10 (64.75\% vs.\ 62.88\%), and 2.08 points on ImageNet-100 (60.69\% vs.\ 58.61\%). The largest improvement is observed on ImageNet-100 at the 20\% ratio, where GLOBE reaches 68.97\%, exceeding the best baseline result of 65.04\% by 3.93 points. Consistent gains are also obtained on SVHN, ImageNet-1K, and CIFAR-100 across all retention ratios, demonstrating that GLOBE is robust to dataset scale.

% We attribute these improvements to the joint effect of trajectory-aware gradient matching and structured sparse optimization. First, gradients provide a criterion that reflects how individual samples affect model optimization. Rather than relying on static representations or a single training snapshot, GLOBE aggregates gradient information across multiple checkpoints, enabling it to identify samples that consistently contribute to the overall training trajectory. This trajectory-level criterion reduces the sensitivity to transient gradients and better captures the long-term optimization value of each sample.

We attribute these improvements to the joint effect of trajectory-aware gradient matching and structured sparse optimization. Instead of relying on static representations or gradients from a single snapshot, GLOBE aggregates sample gradients across multiple checkpoints. This trajectory-level criterion captures samples that consistently influence model optimization, reduces sensitivity to transient gradients, and better reflects their long-term contribution to the full training process. Elastic Net converts trajectory matching into a stable sample-level sparse selection problem, suppressing low-contribution samples while smoothing the weights of correlated candidates. Group LASSO further introduces group-level sparsity to reduce redundant selection among samples with similar gradient information. Together, these regularizers produce a compact and structurally diverse coreset that more accurately reproduces the optimization dynamics of the full dataset.

% Second, the LASSO regularization converts the gradient-matching objective into an explicit sparse selection process. It suppresses samples with limited contribution while retaining a compact support of samples with larger and more persistent weights. Consequently, the selected coreset is not merely representative in the feature space, but is specifically optimized to reproduce the training dynamics of the full dataset. Finally, Group LASSO introduces structured sparsity over groups of similar samples, preventing the optimization from redundantly allocating weights to multiple samples that provide highly overlapping gradient information. By jointly enforcing sample-level and group-level sparsity, GLOBE balances optimization importance, compactness, and structural coverage, which explains its consistent performance gains across datasets and retention ratios.

% \noindent\textbf{Cross-Architecture Generalization.} 

\subsection{Ablation}
\label{sec:ablation}

\begin{table}[t]
\centering
\caption{Ablation results of the main components in GLOBE.}
\label{tab:ablation}
\vspace{-1mm}

\footnotesize
\setlength{\tabcolsep}{5pt}
\renewcommand{\arraystretch}{1}
\setlength{\aboverulesep}{0.25ex}
\setlength{\belowrulesep}{0.25ex}

\begin{tabular}{lc}
\toprule
Method / Configuration & Test Accuracy (\%) \\
\midrule
Full GLOBE            & \textbf{90.21} \\
Final Checkpoint Only & 88.53 \\
GLOBE w/o Group LASSO & 85.32 \\
GLOBE w/o Elastic Net & 81.21 \\
Random Sampling       & 74.31 \\
\bottomrule
\end{tabular}

\vspace{-2mm}
\end{table}

\begin{figure*}
    \centering
    \includegraphics[width=1\linewidth]{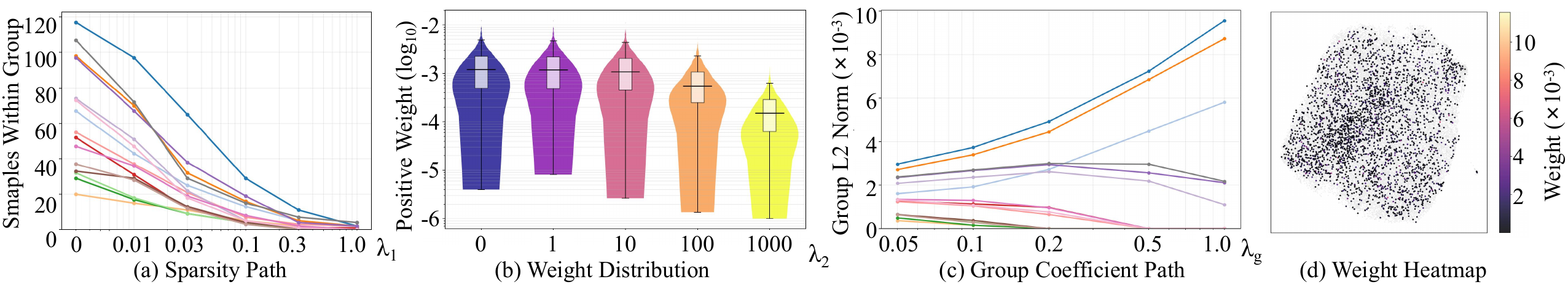}
    \caption{\textbf{Hyperparameter sensitivity and weight analysis of GLOBE.}
(a) Sparsity paths under different values of \(\lambda_1\).
(b) Distribution of positive sample weights as \(\lambda_2\) varies.
(c) Group coefficient paths controlled by \(\lambda_g\).
(d) Heatmap of the learned sample weights in the feature space.}
    \label{fig:sensitivity}
\end{figure*}

\noindent\textbf{Module Ablation.}
We first conduct module-level ablations to evaluate the contribution of each component in GLOBE. All experiments use the same experimental setting (10\% retention ratio, CIFAR-10, ResNet-18), with only the component under investigation removed. Table~\ref{tab:ablation} reports the results.

First, we replace the multi-checkpoint gradient trajectories with gradients extracted only from the final checkpoint to examine the importance of modeling multi-stage training dynamics. Using only the final checkpoint achieves an accuracy of 88.53\%, which is 1.68 percentage points lower than the full model. This result indicates that a single gradient snapshot cannot fully characterize the role of each sample throughout the optimization process, whereas multi-checkpoint gradient trajectories provide a more comprehensive representation of training dynamics.

Second, we remove Group LASSO to evaluate the effect of group-level structured sparsity. Without Group LASSO, the final accuracy decreases from 90.21\% to 85.32\%. This degradation suggests that sample-wise weight optimization alone cannot explicitly exploit the group structure, thereby limiting the ability to model structured redundancy.

Finally, we remove Elastic Net by jointly disabling the $\ell_1$ and squared $\ell_2$ regularization terms. Compared with the full model, the test accuracy decreases by 9 percentage points. Without Elastic Net, the learned weights tend to become either overly dense or locally concentrated, making it more difficult to stably convert the continuous weight solution into a discrete coreset. These components operate at the levels of gradient representation, group-level modeling, and sample-level optimization, respectively, and jointly constitute the complete
GLOBE framework.

% \begin{figure}[t]
%     \centering
%     \includegraphics[width=\linewidth]{figures/module_ablation.pdf}
%     \caption{Ablation results of the key components in GLOBE.
%     \textit{Full} denotes the complete model;
%     \textit{Last Checkpoint Only} uses gradients from only the final
%     checkpoint;
%     \textit{w/o Group LASSO} removes the group-level regularizer; and
%     \textit{w/o Elastic Net} jointly removes the $\ell_1$ and squared
%     $\ell_2$ regularizers. Bar heights indicate the final test accuracy.}
%     \label{fig:module-ablation}
% \end{figure}
\noindent\textbf{Hyperparameter Sensitivity.}
We further conduct a hyperparameter sensitivity analysis under the same experimental setting (10\% retention ratio, CIFAR-10, and ResNet-18), covering $\lambda_1$, $\lambda_2$, and $\lambda_g$. The results are shown in Figure~\ref{fig:sensitivity}.

First, $\lambda_1$ primarily controls sample-level sparsity. As shown in Figure~\ref{fig:sensitivity}(a), increasing $\lambda_1$ drives more small-magnitude weights to zero, leading to a general reduction in the number of nonzero weights across the similarity-based groups formed within each class. The sparsification rates differ across groups, reflecting variations in their gradient-trajectory distributions and levels of sample redundancy. Groups with more homogeneous gradient-trajectory patterns become sparse under relatively small values of $\lambda_1$, whereas groups with greater internal variation require stronger regularization to achieve a comparable sparsity level.

Second, $\lambda_2$ primarily controls the magnitude and distribution of the nonzero weights. As shown in Figure~\ref{fig:sensitivity}(b), small values of $\lambda_2$ result in a heavier upper tail, indicating that the solution assigns relatively large weights to a small number of samples. As $\lambda_2$ increases, these large weights are progressively
suppressed, reducing disparities in weight allocation and shifting the distribution toward smaller magnitudes. This behavior is consistent with the stabilizing effect of squared $\ell_2$ regularization on correlated samples. However, an excessively large $\lambda_2$ may over-shrink the weights, compress their dynamic range, and weaken the distinction among samples.

Finally, $\lambda_g$ controls group-level sparsity. As shown in Figure~\ref{fig:sensitivity}(c), increasing $\lambda_g$ progressively suppresses several groups, with their group-wise $\ell_2$ norms approaching or reaching zero. Meanwhile, the remaining groups retain or increase their norms, indicating a redistribution of weights toward groups that are more important for gradient-trajectory reconstruction. The heterogeneous responses across groups demonstrate that Group LASSO does not uniformly shrink all groups, but instead induces selective group-level sparsity while preserving the groups required to maintain
the reconstruction objective. Figure~\ref{fig:sensitivity}(d) further visualizes the learned sample weights in the feature space. The weights are broadly distributed without pronounced isolated peaks, indicating that the optimization avoids excessive concentration on a few samples. Meanwhile, a substantial proportion of samples retain nonzero weights, preserving sufficient coverage of the data distribution.

\section{Conclusion}
% \noindent We presented GLOBE, a globally optimized coreset selection framework that revisits gradient matching through the lens of dynamic optimization and distributional alignment. By modeling per-sample \emph{gradient trajectories}, GLOBE captures temporal training signals that are overlooked in single-step gradient formulations. Our two-level distribution-matching objective jointly aligns the mean and covariance structure of gradient trajectories, while the combination of Group LASSO, Elastic Net regularization, and simplex
% constraints produces compact and semantically structured coresets. Extensive experiments across diverse benchmarks show that GLOBE achieves consistent and often substantial improvements over state-of-the-art methods, especially under
% aggressive compression. Beyond offering a stronger coreset baseline, our work demonstrates the value of incorporating temporal gradient information and structured sparsity into data-efficient learning pipelines. Future work may extend these ideas to large-scale vision–language models, continual learning, and hardware–algorithm co-design for real-time edge deployment.

\noindent We presented GLOBE, a coreset selection framework that combines multi-checkpoint gradient-trajectory matching with structured sparse optimization. GLOBE represents each sample through its gradient trajectory and jointly preserves the first-order mean and projected second-order moments of the full-data trajectory distribution. Group LASSO and Elastic Net induce group- and sample-level sparsity, while nonnegative budget constraints and class-balanced selection produce compact and representative coresets. Experiments across six image classification benchmarks and five evaluation architectures demonstrate consistent improvements over existing methods, particularly at low retention ratios. These results highlight the effectiveness of combining temporal gradient information with structured sparsity for data-efficient learning.

% \section*{Acknowledgments}

\bibliography{aaai2027}

\end{document}